\documentclass{article}

\usepackage{microtype}
\usepackage{graphicx}
\usepackage{subfigure}
\usepackage{booktabs} 
\usepackage{amsmath}
\usepackage{amsthm}
\usepackage{amssymb,amsfonts}
\usepackage{multirow}
\usepackage{hyperref}

\usepackage[accepted]{icml2020}

\newcommand{\argmin}{\operatornamewithlimits{arg\,min}}

\newtheorem{lemma}{Lemma}
\newtheorem{theorem}{Theorem}

\newtheorem{definition}{Definition}
\newtheorem{remark}{Remark}

\def\WW{\boldsymbol W}
\def\OO{\boldsymbol \Omega}

\def\bb{\boldsymbol b}
\def\ww{\boldsymbol w}
\def\xx{\boldsymbol x}
\def\yy{\boldsymbol y}

\def\oo{\boldsymbol \omega}

\icmltitlerunning{Convolutional Spectral Kernel Learning}

\begin{document}

\twocolumn[
\icmltitle{Convolutional Spectral Kernel Learning}



\icmlsetsymbol{equal}{*}

\begin{icmlauthorlist}
\icmlauthor{Jian Li}{a1,a2}
\icmlauthor{Yong Liu}{a1,a2}
\icmlauthor{Weiping Wang}{a1,a2}
\end{icmlauthorlist}

\icmlaffiliation{a1}{Institute of Information Engineering, Chinese Academy of Sciences}
\icmlaffiliation{a2}{School of Cyber Security, University of Chinese Academy of Sciences}

\icmlcorrespondingauthor{Yong Liu}{liuyong@iie.ac.cn}

\icmlkeywords{Machine Learning, ICML}

\vskip 0.3in
]



\printAffiliationsAndNotice{}  

\begin{abstract}
    Recently, non-stationary spectral kernels have drawn much attention, owing to its powerful feature representation ability in revealing long-range correlations and input-dependent characteristics. However, non-stationary spectral kernels are still shallow models, thus they are deficient to learn both hierarchical features and local interdependence. In this paper, to obtain hierarchical and local knowledge, we build an interpretable convolutional spectral kernel network (\texttt{CSKN}) based on the inverse Fourier transform, where we introduce deep architectures and convolutional filters into non-stationary spectral kernel representations. Moreover, based on Rademacher complexity, we derive the generalization error bounds and introduce two regularizers to improve the performance. Combining the regularizers and recent advancements on random initialization, we finally complete the learning framework of \texttt{CSKN}. Extensive experiments results on real-world datasets validate the effectiveness of the learning framework and coincide with our theoretical findings.
\end{abstract}

\section{Introduction}
With solid theoretical guarantees and complete learning frameworks, kernel methods have achieved great success in various domains over the past decades.
However, compared to neural networks, kernel methods show inferior performance in practical applications because they failed in extracting rich representations for complex latent features.

There are three factors that limit the representation ability of common kernel methods: 
1) Stationary representation \cite{bengio2006curse}. 
Common used kernels are stationary because the kernel function is shift-invariant $\kappa(\xx, \xx') = \kappa(\xx - \xx')$ where the induced feature representations only depend on the distance $\|\xx - \xx'\|$ while free from inputs $\xx$ theirselves.
2) Kernel hyperparameters selection \cite{cortes2010two}.
The assigned hyperparameters of kernel function decide the performance of kernel methods \cite{genton2001classes}. 
Cross-validation (CV) \cite{Cawley2006} and kernel target alignment (KTA) \cite{cortes2010two} were introduced to kernel selection, however, these methods split the process of kernel selection and mode training.
3) Without hierarchical or convolutional architecture.
For example, Gaussian kernels $\kappa(\xx, \xx') = \exp(-\|\xx - \xx'\|/2\sigma^2)$, equivalent to a single layer neural network with infinity width, only characterize the distance $\|\xx - \xx'\|$ and their performance depends on the choice of the kernel hyperparameter $\sigma$.

Yaglom's theorem provides spectral statements for general kernel functions via inverse Fourier transform \cite{yaglom1987correlation}.
To break the limitation of stationary property, non-stationary spectral kernels were proposed with a concise spectral representation based on Yaglom's theorem \cite{samo2015generalized,remes2017non}.
Using Monte Carlo sampling, non-stationary spectral kernels were represented as neural networks \cite{ton2018spatial,sun2019functional} in Gaussian process regression, where kernel hyperparameters can be optimized together with the estimator.
Then, \cite{xue2019deep,li2020automated} extended neural networks of non-stationary spectral kernels to generel learning tasks.
It has been proven that non-stationary kernels can learn both \textit{input-dependent} and \textit{output-dependent} characteristics \cite{li2020automated}.
However, non-stationary kernels fail to extract hierarchical features and local correlations, while deep convolutional neural networks naturally capture those characteristics and present impressive performance \cite{lecun1998gradient,krizhevsky2012imagenet}.

\subsection{Contributions}
In this paper, we propose an effective learning framework (\texttt{CSKN}) which learns rich feature representations and optimize kernel hyperparameters in an end-to-end way. 

{\bf On the Algorithmic Front.}
The framework incorporates non-stationary spectral kernels with deep convolutional neural networks to use the advantages of deep and convolutional architectures. 
Intuitively, the learned feature mapping are \textit{intput-dependent} (non-spectral kernel), \textit{output-dependent} (backpropagation w.r.t. the objective), \textit{hierarchical} (deep architecture) and \textit{local related} (convolutional filters).

{\bf On the Theoretical Front.} 
We derived generalization error bounds of deep spectral kernel networks, revealing how the factors (including architecture, initialization and regularizers) affect the performance and suggesting ways to improve the algorithm.
More importantly, we prove that deeper networks can lead to shaper error bounds with an appropriate initialization schema.
For the first time, we provide a generalization interpretation of the superiority of deep neural networks than relatively shallow networks.

\subsection{Related Work}
Based on Bochner's theorem, the first approximate spectral representations were proposed for shift-invariant kernels \cite{rahimi2007random}, known as random Fourier features.
In theory, \cite{bach2017EquivalenceKernelQuadrature,rudi2017generalization} provided the optimal learning guarantees for random features.
Stacked random Fourier features as neural networks were presented in \cite{zhang2017stacked}.
Based on Yalom's theorem, \cite{samo2015generalized} provided general spectral representations for arbitrary continuous kernels.
Spectral kernel networks have attracted much attention in Gaussian process \cite{remes2017non,sun2018differentiable} and were extended to general learning domains \cite{xue2019deep,li2020automated}.

Deep convolutional neural networks (CNNs) have achieved unprecedented accuracies on in domains including computer vision \cite{lecun1998gradient,krizhevsky2012imagenet} and nature language processing \cite{kim2014convolutional}.
Convolutional neural networks were encoded in a reproducing kernel Hilbert space (RKHS) to obtain invariance to particular transformations \cite{mairal2014convolutional} in an unsupervised fashion. Then, combined with Nystr\"om method, convolutional kernel networks were proposed in an end-to-end manner \cite{mairal2016end}, while its stability to deformation was studied in \cite{bietti2017invariance,bietti2019group}.
Except for stability theory, group invariance was also learned \cite{mallat2012group,wiatowski2017mathematical}. 
Recent research also explored the approximate theory of CNNs via downsampling \cite{zhou2020theory} and universality of CNNs \cite{zhou2020universality}.
Besides, \cite{shen2019learning} introduced convolutional filters to spectral kernels and studied the len of spectrograms.

However, the generalization ability of spectral kernel networks was rarely studied. 
Using Rademacher complexity, the generalization ability of spectral kernels was studied in \cite{li2020automated}.
The RKHS norm and spectral norm were considered to improve the generalization ability of neural networks \cite{bartlett2017spectrally,belkin2018understand,bietti2019kernel}.
Furthermore, \cite{allen2019learning,arora2019fine} proposed that the learnability of deep modes involves both generalization ability and trainability.
Based on the mean field theory, \cite{poole2016exponential,schoenholz2017deep} revealed that initialization schema determines both the trainability and the expressivity.

\section{Preliminaries}
Consider a supervised learning scenario where training samples $D=\{\xx_i, \yy_i\}_{i=1}^n$ are drawn i.i.d. from a fixed but unknown distribution $\rho = \mathcal{X} \times \mathcal{Y}.$ 
Specifically, for general machine learning tasks, we assume the input space be $\mathcal{X} = \mathbb{R}^{d_0}$ and the output space be $\mathcal{Y} \subseteq \mathbb{R}^K,$ where $K=1$ for univariable labels (binary or regression) and $K > 1$ for multivariable labels (multi-class or multi-labels). 

Kernel methods include mappings from the input space $\mathcal{X}$ to a reproducing kernel Hilbert space (RKHS) $\mathcal{H}$ via an implicit feature mapping $\phi: \mathcal{X} \to \mathcal{H},$ which is induced by a Mercer kernel $\kappa(\xx, \xx') = \langle \phi(\xx), \phi(\xx') \rangle$. 
Classical kernel methods learn the prediction function $f: \mathcal{X} \to \mathcal{Y}$ which learns modes in the RKHS space, admitting the linear form $f(x) = \langle \WW, \phi(\xx)\rangle_\mathcal{H}.$
The hypothesis space is denoted by
\begin{align*}
    H_\kappa = \Big\{f | \xx \to f(x) = \langle \WW, \phi(\xx)\rangle_\mathcal{H}\Big\},
\end{align*}
where $\WW \in \mathcal{H} \times \mathcal{Y}$ is the weight of the estimator and the feature mapping $\phi: \mathcal{X} \to \mathcal{H}$ is from the input space to a latent space to characterize more powerful feature representations.
The goal of supervised learning is to learn an ideal estimator $f(\xx)$ to minimize the expected loss
\begin{align}
    \label{equation.expected-loss}
    \inf_{f \in H_\kappa}, ~ \mathcal{E}(f) = \int_{\mathcal{X} \times \mathcal{Y}} \ell(f(\xx), \yy) d \rho(\xx, \yy),
\end{align}
where $\ell$ is the loss function associated to specific tasks.

\subsection{Shift-invariant Kernels}
Shift-invariant kernels only depend on the distance $\tau = \xx - \xx'$, written as $\kappa(\xx, \xx') = \kappa(\tau).$ 
Commonly used kernels are shift-invariant (stationary), such as Gaussian kernels $\kappa(\xx, \xx') = \exp(-\|\tau\|_2^2)$ and Laplacian kernels $\kappa(\xx, \xx') = \exp(-\|\tau\|_1)$.
According to Bochner's theorem, shift-invariant kernels are determined by its spectral density $s(\oo)$ via inverse Fourier transform \cite{stein2012interpolation}.
\begin{lemma}[Bochner's theorem]
    A shift-invariant kernel $\kappa(\xx, \xx') = \kappa(\xx - \xx')$ on $\mathcal{X}$ is positive
    definite if and only if it can be represented as
    \begin{align}
        \label{equation.bochner}
        \kappa(\xx, \xx') & = \int_{\mathcal{X}} e^{i \oo^T(\xx - \xx')} s(\oo) d \oo,
    \end{align}
    where $s(\oo)$ is a non-negative probability density.
\end{lemma}

Based on Bochner's theorem \eqref{equation.bochner} and Monte Carlo sampling, random Fourier features were proposed to approximate shift-invariant kernels via $\kappa(\xx, \xx') \approx \langle \psi(\xx), \psi(\xx') \rangle$ \cite{rahimi2007random}:
\begin{align}
    \label{equation.shift-invariant-rf}
    \psi(\xx) = \sqrt{\frac{2}{D}} \cos(\OO^T\xx + \bb),
\end{align}
where the frequency matrix $\OO = \{\oo_1, \oo_2, \cdots, \oo_D\}$ is drawn from the spectral density $s(\oo)$ and the phase vector $\bb = \{b_1, b_2, \cdots, b_D\}$ is drawn uniformly from $[0, 2\pi]^D$.

\subsection{Non-stationary Spectral Kernels}
Shift-invariant kernels $\kappa(\tau) = \kappa(\xx - \xx')$ are stationary, which only take into account the distance $\xx - \xx'$ but neglect useful information of the inputs themselves, also called stationary spectral kernels.
However, the most general family of kernels are non-stationary, i.e. linear kernels $\kappa(\xx, \xx') = \xx^T\xx'$ and polynomial kernels $\kappa(\xx, \xx') = (\xx^T\xx' + 1)^r$.

Recently, based on Yaglom's theorem, the Fourier analysis theory has been extended to general kernels, including both stationary and non-stationary cases \cite{samo2015generalized}.

\begin{lemma}[Yaglom's theorem]
    A general kernel $\kappa(\xx, \xx')$ is positive definite on $\mathcal{X}$ is positive define if and only if it admits the form
    \begin{align}
        \label{equation.yaglom}
        \kappa(\xx, \xx') & = \int_{\mathcal{X} \times \mathcal{X}} e^{i (\oo^T\xx - \oo'^T\xx')} \mu (d \oo, d \oo'),
    \end{align}
    where $\mu (d \oo, d \oo')$ is a Lebesgue-Stieltjes measure associated to some positive semi-definite (PSD) spectral density function $s(\oo, \oo')$ with bounded variations.
\end{lemma}

Yaglom's theorem illustrates that a general kernel $\kappa(\xx, \xx')$ is associated to some positive semi-definite spectral density $s(\oo, \oo')$ over frequencies $\oo, \oo'.$
Meanwhile, shift-invariant kernels (Bochner's theorem) is a special case of spectral kernels (Yaglom's theorem) when the spectral measure is concentrated on the diagonal $\oo = \oo'.$

To ensure a valid positive semi-definite spectral density in \eqref{equation.yaglom}, we symmetrize spectral densities where $s(\oo, \oo') = s(\oo', \oo)$ and then introduce the diagonal components $s(\oo, \oo), s(\oo', \oo')$ \cite{samo2015generalized,remes2017non}, such that the kernel is defined as
\begin{align}
    \label{equation.non-stationary-kernel}
    \kappa(\xx, \xx') = \int_{\mathcal{X} \times \mathcal{X}} \mathcal{E}_{\oo, \oo'}(\xx, \xx') \mu (d \oo, d \oo')
\end{align}
where the exponential term is 
\begin{align*}
    \mathcal{E}_{\oo, \oo'}(\xx, \xx')
    ~=~ &\frac{1}{4} \Big[
    e^{i (\oo^T\xx - \oo'^T\xx')} ~+~ e^{i (\oo'^T\xx - \oo^T\xx')} \\
    &~+ ~e^{i \oo^T(\xx - \xx')} ~+~ e^{i \oo'^T(\xx - \xx')} \Big].
\end{align*}

Similar to the approximation of shift-invariant kernels \eqref{equation.shift-invariant-rf}, we derive a finite-dimensional approximation of non-stationary kernels \eqref{equation.non-stationary-kernel} by performing Monte Carlo method
\begin{align*}
    \kappa(\xx, \xx') \approx \langle \psi(\xx), \psi(\xx) \rangle.
\end{align*}
The random Fourier features for non-stationary kernels are
\begin{align}
    \label{equation.non-startionary-rff}
    \psi(\xx) = \frac{1}{\sqrt{2D}}
    \begin{bmatrix}
        \cos(\OO^T \xx + \bb) + \cos(\OO'^T \xx + \bb')
    \end{bmatrix},
\end{align}
where the frequency matrices $\OO, \OO' \in \mathbb{R}^{d_0 \times D}$ are paired Monte Carlo samples, $\OO = \{\oo_1, \oo_2, \cdots, \oo_D\}, \OO' = \{\oo'_1, \oo'_2, \cdots, \oo'_D\} ,$ the frequency pairs $\{(\oo_i, \oo'_i)\}_{i=1}^D \in \mathbb{R}^{d_0}$ are drawn i.i.d. from the spectral density $s(\oo, \oo')$.
The phase vectors $\bb$ and $\bb'$ are drawn uniformly from $[0, 2\pi]^D.$

\section{Convolutional Spectral Kernel Learning}

\begin{figure*}
    \begin{center}
        \includegraphics[width=.9\textwidth]{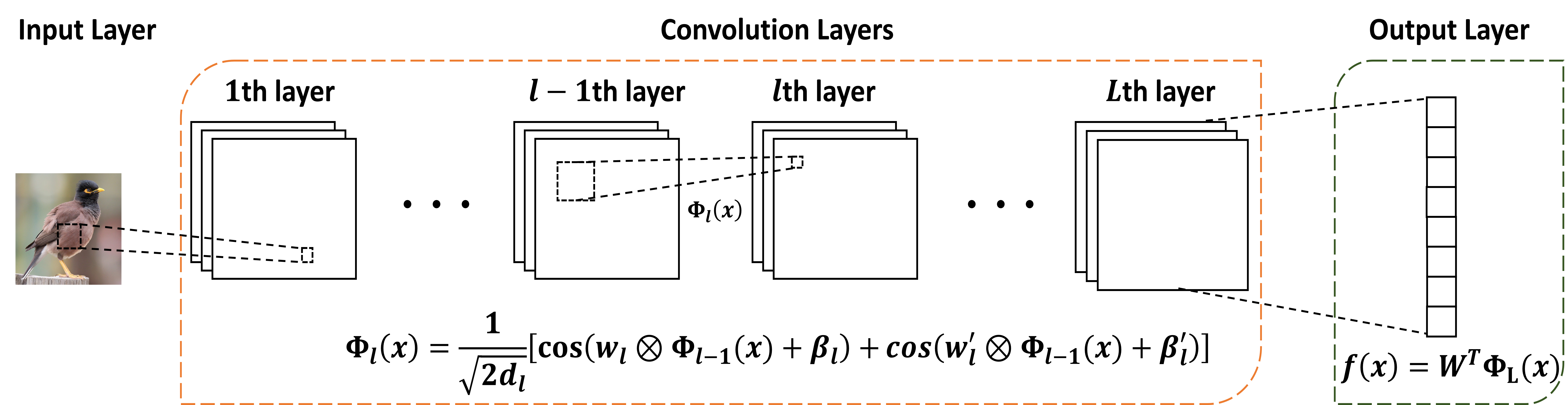}
    \end{center}
    \caption{The structure of the learning framework}
    \label{fig.structure}
\end{figure*}

\subsection{Multilayer Spectral Kernel Networks}
In the view of neural networks, a non-stationary kernel \eqref{equation.non-stationary-kernel} is a single-layer neural network with infinite width, while the random Fourier approximation \eqref{equation.non-startionary-rff} reduce the infinite dimension to a finite width.
Even though the non-stationary kernels characterize input-dependent features, it is deficient in feature representations due to its shallow architecture.

In this paper, we use the deep architectures of non-stationary spectral kernels by stacking their random Fourier features in a hierarchical composite way:
\begin{equation*}
    \begin{aligned}
        \kappa(\xx, \xx') &\approx \langle \Psi_L(\xx), \Psi_L(\xx') \rangle \qquad \text{with}\\
        \Psi_L(\xx) &= \psi_L(\cdots \psi_2(\psi_1(\xx)))
    \end{aligned}    
\end{equation*}
where the kernel $\kappa$ consists of $L$-layers stacked spectral kernels and the feature mappings for any layer are approximated by random Fourier features \eqref{equation.non-startionary-rff}.
Based on the feature mapping of the last layer $\Psi_{l-1}(\xx),$ we explicitly definite the random Fourier mapping of $l$-th layer $\Psi_{l}: \mathbb{R}^{d_{l-1}} \to \mathbb{R}^{d_l}, l = 1, 2, \cdots, L$
\begin{align*}
    \Psi_{l}(\xx)  = 
    \frac{1}{\sqrt{2D}}
    \Big[
        &\cos(\OO_l^T \Psi_{l-1}(\xx) + \bb_l) \\
        +~ &\cos(\OO_l'^T \Psi_{l-1}(\xx) + \bb_l')
    \Big],
\end{align*}
where $\Psi_0(\xx) = \xx$ is the input data, the frequency pairs in the $l$-th frequency matrices $\OO_l, \OO'_l$ are drawn i.i.d. from the $l$-th layer's spectral density $s_l(\oo, \oo')$. The elements in $l$-th phase vector $\bb_l$ are drawn uniformly from $[0, 2\pi]^{d_l}.$

The above architecture of deep spectral kernel networks is a kind of fully connected network (FCN), where the network includes $L$ convolutional layers and two frequency matrices $\OO_l, \OO'_l \in \mathbb{R}^{d_l \times d_{l-1}}$ and two bias vectors $\bb_l, \bb'_l \in \mathbb{R}^{d_l}$ for the $l$-th layer.
Therefore, the $l$-th convolutional layer involves $2 \times d_l \times (d_{l-1} + 1)$ parameters.

\subsection{Convolutional Spectral Kernel Networks}
Even though the multilayer spectral kernel representations can learn input-dependent characteristics, long-range relationships and hierarchical features, this full connected network (FCN) fails to extract local correlations on the structural dataset, i.e. image and natural language.
However, convolutional networks guarantee the local connectivity, promising dramatic improvements in complex applications.

For the sake of simplicity, 
we integrate spectral kernel networks with convolutional architecture but without pooling layers and skip connections.
We define the convolutional spectral kernel network (\texttt{CSKN}) in a hierarchical kernel form by stacking spectral kernels
$
        \kappa(\xx, \xx') \approx \langle \Phi_L(\xx), \Phi_L(\xx') \rangle.
$

For each channel of the $l$-th convolutional layer, the convolutional mapping $\Phi_l: \mathbb{R}^{d_{l-1}} \to \mathbb{R}^{d_l}$ is defined as
\begin{equation}
    \begin{aligned}
        \label{equatiion.convolution-feature-mapping}
        \Phi_{l}(\xx)  = 
        \frac{1}{\sqrt{2d_l}}
        \Big[
            &\cos(\ww_l \ \otimes \Phi_{l-1} (\xx) + \beta_l) \\
            +~ &\cos(\ww_l' \otimes \Phi_{l-1}(\xx) + \beta_l')
        \Big], 
    \end{aligned}
\end{equation}
where $l=1, \cdots, L,$ $\Phi_0(\xx) = \xx$ and the $l$-th convolutional filters are pairwise $\ww_l, \ww'_l \in \mathbb{R}^{d'_l}$ in the filter size $d'_l$.
The frequency pair $(\ww_l, \ww'_l)$ is drawn from the spectral density $s(\ww_l, \ww_l')$ for $l$-th layer convolutional spectral kernel. The bias terms $\beta_l, \beta'_l$ are uniformly sampled from $[0, 2\pi].$

We assume there is $c_l$ channels for the $l$-th convolutional layer.
Due to weights sharing, the $l$-th layer exists $c_{l-1} \times c_l$ convolutional feature mappings $\Phi_l(\xx)$ in \eqref{equatiion.convolution-feature-mapping}. 
Thus, there are $2 \times (d'_l + 1) \times c_{l-1} \times c_l$ parameters for the $l$-th convolutional layer, because $c_{l-1}$ and $c_l$ are small constants thus the number of parameters is also dramatically reduced. 

\subsection{Learning Framework}
The structure of estimator $f(\xx)$ is shown as Figure \ref{fig.structure}.
Because it is hard to estimate the minimization of the expected loss (\ref{equation.expected-loss}), so we aim to minimize the empirical loss.

Based on theoretical findings (Theorem \ref{thm.generalization-DSKL} in next section), we incorporate the empirical loss with two kinds of regularization terms in the minimization objective, written as 
\begin{equation}
    \label{equation.primal-objective}
    \resizebox{0.9\linewidth}{!}{$
    \displaystyle
    \begin{split}
        \argmin_{\WW, \Phi_L} \underbrace{\frac{1}{n} \sum_{i=1}^n \ell(f(\xx_i), \yy_i)}_{g(\WW)} + \lambda_1 \|\WW\|_* + \lambda_2 \|\Phi_L(\boldsymbol{X})\|_F^2
    \end{split}
    $}
\end{equation}
where the depth is $L$ and $l \in [1, \cdots, L]$. 
The estimator is $f(\xx_i)=\WW^T\Phi_L(\xx_i) \in \mathbb{R}^K$, where the weighted matrix is $\WW \in \mathbb{R}^{d_L \times K}$ and we employ the deep convolutional spectral kernel representations $\Phi_L: \mathbb{R}^{d_0} \to \mathbb{R}^{d_L}$ in  a hierarchical composite way \eqref{equatiion.convolution-feature-mapping}. 
The trace norm $\|\WW\|_*$ regularize the estimator weights and the squared Frobenius norm $\|\Phi_L(\boldsymbol{X})\|_F^2 = \sum_{i=1}^n \|\Phi_L(\xx_i)\|^2_F$ is used to regularize the feature mappings on all samples.
These two norms are scarcely used in conventional methods, where $\|\WW\|_*$ represents the RKHS norm of primal kernel methods and $\|\Phi_L(\boldsymbol{X})\|_F^2$ regularizes the frequency pairs $(\ww_l, \ww'_l)$.

Using backpropagation w.r.t the objective, we update the model weights $\WW$ and frequency pairs $(\ww_l, \ww_l')$ for convolutional layers in the objective \eqref{equation.primal-objective}, that makes the feature mappings $\Phi_l(\xx)$ dependent on the specific tasks.
The spectral density $s(\ww_l, \ww'_l)$, the key of kernel methods' generalization ability, is modified via
the update of frequency pairs $(\ww_l, \ww_l')$, where kernel hyperparameters in the spectral densities are optimized in an end-to-end manner.

\subsection{Update $\WW$ via Singular Value Thresholding (SVT)}
The updates of gradient of $\WW$ involves trace norm in (\ref{equation.primal-objective}), but we can't update $\WW$ using gradient descent methods because the trace norm is nondifferentiable.
So, we employ singular value thresholding (SVT) \cite{cai2010singular} to solve the minimization of trace norm in the two steps:\\
1) Update $\WW$ with SGD on the empirical loss 
    \begin{align*}
        {\boldsymbol Q} = \WW^{t}-\eta\nabla g(\WW^{t}),
    \end{align*}
    where $\eta$ is the learning rate and ${\boldsymbol Q}$ is an intermediate.\\
2) Update $\WW$ with SVT on the trace norm 
\begin{align*}
    \WW^{t+1} = {\boldsymbol U}\text{diag}\big(\left\{\sigma_j - \lambda_1\eta\right\}_+\big){\boldsymbol V}^T,
\end{align*}
where ${\boldsymbol Q}={\boldsymbol U}{\boldsymbol \Sigma}{\boldsymbol V}^T$ is the singular values decomposition, ${\boldsymbol \Sigma}$ is the diagonal ${\text{diag}(\{\sigma_j\}_{1 \leq i \leq r})}$ and $r$ is the rank of ${\boldsymbol Q}$.

\subsection{Random Initialization}
To approximate non-stationary kernels, we use random Gaussian weights as initialization.
We initialize the joint probability distribution $s(\ww_l, \ww_l')$ for the $l$-th layer as two independent normalization distributions with zero mean and the variance $\sigma_l$ for all dimensions
\begin{align}
    \label{def.initial-spectral-density}
    [\ww_l]_{i} \sim \mathcal{N}(0, \sigma_l^2), \quad
    [\ww'_l]_{i} \sim \mathcal{N}(0, \sigma_l^2) 
\end{align}
where $i = 1, \cdots, d'_l.$
According to mean field theory, we select the Gaussian initialization hyperparameters $\sigma_l$ for every layer to achieve the critical line between order-to-chaos transition and satisfy dynamical isometry \cite{poole2016exponential,pennington2017resurrecting}.

\section{Generalization Analysis}
Rademacher complexity theory has achieved great success in shallow learning, however it's an open problem whether Rademacher complexity is applicative for deep neural networks \cite{belkin2018understand,bietti2019on}.
In this section, we apply Rademacher complexity theory to spectral kernel networks and explore how the factors in \texttt{CSKN} affect the generalization performance.

Firstly, we derive the generic generalization error bounds for kernel methods based on Rademacher complexity.
The empirical Rademacher complexity is mainly dependent on the sum of diagonals $\kappa(\xx_i, \xx_i'), i = 1, \cdots, n.$
So, we explore the generalization error bounds of three different architectures: 1) shift-invariant kernels, 2) non-stationary spectral kernels, 3) deep non-stationary spectral networks.
We then discuss the approximation ability of random Fourier features and the use of convolutional filters.

\begin{definition}
    The empirical Rademacher complexity of hypothesis space $H_\kappa$ is defined as
    \begin{align*}
         & \widehat{\mathcal{R}}(H_\kappa)
        = \frac{1}{n} ~ \mathbb{E}_\xi \left[\sup_{f \in H_\kappa} \sum_{i=1}^{n} \sum_{k=1}^K \xi_{ik} [f(\xx_i)]_k\right],
    \end{align*}
    where $[f(\xx_i)]_k$ means the $k$-th value of the outputs and $\xi_{ik}$s are $n \times K$ independent Rademacher variables.
    The expected Rademacher complexity is $\mathcal{R}(H_\kappa)=\mathbb{E} ~ \widehat{\mathcal{R}}(H_\kappa)$.
\end{definition}

\subsection{Excess risk bound for kernel methods}
\begin{lemma}
    \label{lem.excess_risk_bounds}
    Assume the loss function $\ell$ is $L$-Lipschitz for $\mathbb{R}^K$ equaipped with the $2$-norm.
    With a probability at least $1-\delta$, the excess risk bound holds
    \begin{align*}
        \mathcal{E}(\widehat{f}_n) - \mathcal{E}(f^*) \leq 4 \sqrt{2} L \widehat{\mathcal{R}}(H_\kappa) + \mathcal{O}\Big(\sqrt{\frac{\log 1/\delta}{n}}\Big),
    \end{align*}
    where $f^* \in H_\kappa$ is the most accurate estimator in the hypothesis space, $\widehat{f}_n$ is the empirical estimator.
    The empirical Rademacher complexity $\widehat{\mathcal{R}}$ is bounded by
        \begin{equation}
            \label{inequation.grc-bound-result-rff}
            \begin{aligned}
                 \widehat{\mathcal{R}}(H_\kappa)
                \leq ~ \frac{B}{n} \sqrt{K \sum_{i=1}^n \kappa(\xx_i, \xx_i)}
            \end{aligned}
        \end{equation}
    where the upper bound of the trace norm on $\WW$ is $B = \sup_{f \in H_\kappa} \|\WW\|_* < \infty$.
\end{lemma}

Based on Rademacher complexity, generalization error bounds of kernel methods have been well-studied \cite{bartlett2002rademacher,cortes2013learning}, where the convergence depends on empirical Rademacher complexity $\widehat{\mathcal{R}}(H_\kappa)$. 
Meanwhile, empirical Rademacher complexity is determined by the trace of empirical kernel matrix $\sum_{i=1}^n \kappa(\xx_i, \xx_i)$.
The upper bound of Rademacher complexity is related to the corresponding kernel function.

\begin{remark}
    From \eqref{inequation.grc-bound-result-rff}, we find that the minimization of Rademacher complexity need to minimize both $B$ and the sum of diagonals $\sum_{i=1}^n \kappa(\xx_i, \xx_i)$.
    Because $B$ is the upper bound of the trace norm $\|\WW\|*$ and the trace holds $\sum_{i=1}^n \langle \phi(\xx_i), \phi(\xx_i) \rangle = \sum_{i=1}^n \|\phi(\xx_i)\|_2^2 = \|\phi(\boldsymbol{X})\|_F^2,$ we introduce $\|\WW\|*$ and $\|\phi(\boldsymbol{X})\|_F^2$ as regularizers to obtain better performance, leading to the objective in \eqref{equation.primal-objective}.
\end{remark}

\subsection{Rademacher Complexity of Shift-invariant Kernels}
According to the Bochner's theorem \eqref{equation.bochner}, we define shift-invariant kernels as $\kappa(\xx, \xx') = \mathbb{E}_{\oo \sim s(\oo)} ~ \cos[\oo^T (\xx_i - \xx'_i)].$
\begin{lemma}
    \label{thm.rc-shift-invariant-kernel}
    For arbitrary shift-invariant kernels, the diagonal element of the corresponding kernel matrix is 
    \begin{align*}
        \kappa(\xx_i, \xx_i) = \mathbb{E}_{\oo \sim s(\oo)} ~ \cos[\oo^T (\xx_i - \xx_i)] = 1.
    \end{align*}
     where $i=1, \cdots, n.$ 
\end{lemma}

For shift-invariant kernels, diagonals of shift-invariant kernels identically equal to one regardless of the spectral density $s(\oo).$
The trace equal to $\sum_{i=1}^n \kappa(\xx_i, \xx_i) = n$.
The convergence rate of Rademacher complexity is $\widehat{\mathcal{R}}(H_\kappa) \leq \mathcal{O}(\sqrt{K/n})$ when we bound the norm $\|\WW\|_* \leq c$ with a constant $c$ \cite{bartlett2002rademacher}.

\subsection{Improvements of Non-stationary Spectral Kernels}
Based on the Yaglom's theorem \eqref{equation.yaglom}, we define the non-stationary spectral kernels $\kappa(\xx, \xx')$ as 
\begin{equation*}
    \begin{split}
    ~ \mathbb{E}_{\oo,\oo'} ~ \frac{1}{4} \Big[
        &\cos(\oo^T\xx - \oo'^T\xx') 
    ~+~ \cos (\oo'^T\xx - \oo^T\xx') \\
    ~+~ &\cos (\oo^T\xx - \oo^T\xx') ~+~ \cos (\oo'^T\xx - \oo'^T\xx') \Big].
    \end{split}
\end{equation*}
where the frequency pair $\oo, \oo'$ is drawn i.i.d from the spectral density $s(\oo, \oo').$
We initialize the spectral density $s(\oo, \oo')$ as two independent Gaussian distributions $\oo \sim \mathcal{N}(0, \sigma^2)$ and $\oo' \sim \mathcal{N}(0, \sigma^2)$, where $\sigma > 0.$

\begin{theorem}
\label{thm.rc-spectral-kernel}
The diagonals of a non-stationary spectral kernel matrix are:
\begin{equation*}
    \begin{aligned}
        \kappa(\xx_i, \xx_i) 
        &= \mathbb{E}_{\oo, \oo'} ~ \frac{1}{2}\Big[\cos\big[(\oo - \oo')^T \xx_i\big] + 1\Big] \\
        &= \frac{1}{2} \left[\exp\left(-\sigma^2 \xx_i^T\xx_i \right) + 1 \right]
    \end{aligned}
\end{equation*}
where the frequencies are drawn from the joint spectral density $\oo, \oo' \sim s(\oo, \oo')$.
\end{theorem}

Due to $\sigma > 0$ and $\xx_i^T\xx_i > 0$, the diagonal elements of non-stationary spectral kernels are less than $1$.
When the variance $\sigma$ is large, the trace $\sum_{i=1}^n \kappa(\xx_i, \xx_i)$ is even smaller than the case in stationary kernels (shift-invariant kernels).
Note that, shift-invariant kernels are the special case of spectral kernels with the diagonal density $\oo = \oo',$ where all diagonals are $\kappa(\xx_i, \xx_i') = 1$ for $i=1, 2, \cdots, n.$

\subsection{Improvements from Deep Architecture}
We introduce deep architecture for spectral kernels via 
\begin{equation*}
        \kappa_L(\xx, \xx')
        =~ \langle \phi_L(\xx), \phi_L(\xx') \rangle,
\end{equation*}
where the $l$-th layer spectral representations of stacked spectral kernels is related to its last layer in a recursive way:
\begin{equation*}
    \begin{split}
        &\langle \phi_l(\xx), \phi_l(\xx') \rangle =  \\
    \mathbb{E}_{\oo_l,\oo_l'} ~ \frac{1}{4} \Big[ &\cos(\oo_l^T\phi_{l-1}(\xx) - \oo_l'^T\phi_{l-1}(\xx')) \\
    ~+~ &\cos (\oo_l'^T\phi_{l-1}(\xx) - \oo_l^T\phi_{l-1}(\xx')) \\
    ~+~ &\cos (\oo_l^T\phi_{l-1}(\xx) - \oo_l^T\phi_{l-1}(\xx')) \\
    ~+~ &\cos (\oo_l'^T\phi_{l-1}(\xx) - \oo_l'^T\phi_{l-1}(\xx')) \Big],
    \end{split}
\end{equation*}
where $\phi_0(\xx) = \xx$ represents the inputs and $\kappa_0(\xx, \xx') = \xx^T\xx'$.
We use a simple initialization schema where the paired frequencies $(\oo_l, \oo'_l)$ are drawn i.i.d. from two independent Gaussian distributions with $\oo_l \sim \mathcal{N}(0, \sigma)$ and $\oo'_l \sim \mathcal{N}(0, \sigma).$

\begin{theorem}
    \label{thm.generalization-DSKL}
    For any input data $\xx_i$, the diagonal $\kappa_l(\xx_i, \xx'_i)$ is smaller than the diagonal $\kappa_{l-1}(\xx_i, \xx'_i)$ of laster layer:
    \begin{equation*}    
        \begin{aligned}
            &\kappa_l(\xx_i, \xx_i)  \\
            =~ &\mathbb{E}_{\oo_l, \oo_l'} ~ \frac{1}{2}\Big[\cos\big[(\oo_l - \oo_l')^T \phi_{l-1}(\xx_i)\big] + 1\Big] \\
            =~ &\frac{1}{2} \left[\exp\left(-\sigma_l^2 \kappa_{l-1}(\xx_i, \xx_i) \right) + 1 \right]\\
            \leq~ &\kappa_{l-1}(\xx_i, \xx_i),
        \end{aligned}
    \end{equation*}
    when the variance $\sigma_l^2$ satisfy 
    \begin{align}
        \label{inequality.variance}
        \sigma_l^2 \geq - \frac{\log ~ [2\kappa_{l-1}(\xx_i, \xx_i) -1]}{\kappa_{l-1}(\xx_i, \xx_i)}.
    \end{align}
\end{theorem}

\begin{remark}
    Theorem \ref{thm.generalization-DSKL} holds for all diagonals $\kappa_l(\xx_i, \xx_i) \leq \kappa_{l-1}(\xx_i, \xx_i)$, thus the sum of diagonals magnify the difference. 
    With favorable initialization schema, we obtain decreasing diagonals as the depth increases, which leads to sharper generalization error bounds.
    It's worth noting that, for the first time, {\bf we prove deeper architectures of neural networks can obtain better generalization performance} with suitable initialization.
    The theorem reveals the superiority of deep neural networks than shallow learning (such as kernel methods) in the view of generalization.
\end{remark}

The results in Theorem \ref{thm.generalization-DSKL} guide the design of the variance $\sigma_l$ to get better generalization performance for deep neural networks.
The right of inequality \eqref{inequality.variance} has decreasing property w.r.t. the diagonals $\kappa_{l-1}(\xx_i, \xx_i)$.
To make deeper architecture available, we should ensure the decreasing on the diagonals $\kappa_l(\xx_i, \xx_i)$ w.r.t. the depth $l$, such that we enlarge $\sigma_l$ for the increasing depth $l$.
Based on the mean field theory, recent work has devised the better initialization strategies \cite{poole2016exponential,yang2017mean,hanin2018start,jia2019orthogonal} to improve the trainability, however these strategies are irrelevant to the depth, ignoring the issues in generalization.
It's worthy to further study the initialization schema which characterizes both good generalization ability and trainability.

\subsection{Trainable Spectral Kernel Network}
We derive above generalization analysis in Lemma \ref{thm.rc-shift-invariant-kernel}, Theorem \ref{thm.rc-spectral-kernel}, Theorem \ref{thm.generalization-DSKL} in the RKHS space with implicit feature mappings.
However, the computation of hierarchical stacked spectral kernels is intractable and optimal kernel hyperparameters are hard to estimate, so we construct explicit feature mappings via Monte Carlo approximation in \eqref{equation.non-startionary-rff}, where $\kappa(\xx, \xx') \approx \langle \psi(\xx), \psi(\xx') \rangle$ and $\psi: \mathbb{R}^{d_0} \to \mathbb{R}^D$. 

According to Hoeffding’s inequality, we bound the approximation error with a probability of at least $1 - \eta$:
\begin{align*}
    |\langle \psi(\xx), \psi(\xx') \rangle - \kappa(\xx, \xx')| \leq \sqrt{\frac{2}{D}\log\frac{2}{\eta}}.
\end{align*}
where $\eta \in (0, 1)$ is a small constant. 
The approximation error converges fast with the number of Monte Carlo samplings $D$.
\cite{rahimi2007random} has proven small approximate error $\epsilon$ is achieved by any constant probability when $D = \Omega(\frac{d_0}{\epsilon^2}\log\frac{1}{\epsilon})$.
Besides, recent work revealed $D = \mathcal{O}(\sqrt{n})$ random features can achieve optimal learning rates in kernel ridge regression tasks \cite{rudi2017generalization,bach2017EquivalenceKernelQuadrature}. 

Traditional kernel selection methods split the choice of hyperparameters and model learning.
In contrast, the presented spectral kernel networks are trainable, thus we can optimize the kernel hyperparameters and model weights together, which are trained in an end-to-end manner.

\subsection{The Use of Convolutional Filters}
Deep convolutional neural networks \cite{lecun1998gradient,krizhevsky2012imagenet} have achieved impressive accuracies which are often attributed to effectively leverage of the local stationarity of natural images at multiple scales.
The group invariance and stability to the action of diffeomorphisms were well-studied in \cite{mallat2012group,wiatowski2017mathematical,bietti2019group}.
Meanwhile, \cite{zhou2020universality} studied the universality of deep convolutional neural networks and proved that CNNs can be used to approximate any continuous function to an arbitrary accuracy when the depth is large enough.
However, the generalization ability of CNN was scarcely studied, because it's hard to extend the generalization results of FCN to CNN due to different structures.
It's not clear how to prove the superiority of convolutional networks in the view of generalization.

\begin{table*}[t]
    \centering
    \begin{tabular}{@{\extracolsep{0.8cm}}l|ccc|cc}
        \toprule
         Dataset           & CNN                         & CRFF                        & DSKN                       & CDSK                                   & \texttt{CSKN}             \\ \hline
         segment           & 95.24$\pm$1.72              & 95.35$\pm$2.17              & 96.08$\pm$1.94             & 96.37$\pm$1.21                         & \textbf{97.03$\pm$1.42}   \\
         satimage          & 86.74$\pm$1.49              & 85.46$\pm$1.86              & 86.56$\pm$1.80             & \underline{88.31$\pm$1.37}             & \textbf{88.35$\pm$1.25}   \\
         usps              & 97.81$\pm$1.74              & 97.76$\pm$2.03              & \underline{99.14$\pm$1.64}             & 98.17$\pm$1.56                         & \textbf{99.18$\pm$1.27}   \\
         pendigits         & 99.07$\pm$0.57              & 99.03$\pm$0.67              & 99.16$\pm$0.50             & \underline{99.44$\pm$0.57}             & \textbf{99.46$\pm$0.41}   \\
         letter            & 95.70$\pm$1.47              & 95.34$\pm$1.56              & 96.16$\pm$1.71             & 96.69$\pm$1.46                         & \textbf{96.97$\pm$1.31}   \\
        \bottomrule
    \end{tabular}
    \caption{
        \normalsize  Classification accuracy (\%) for all datasets. We bold the numbers of the best method and underline the numbers of the other methods which are not significantly worse than the best one. }
    \label{tab.small-dataset}
\end{table*}

\begin{table*}[t]
    \centering
    \begin{tabular}{@{\extracolsep{0.8cm}}l|ccc|cc}
        \toprule
         Train size           & CNN             & CRFF            & DSKN                        & CDSK                       & \texttt{CSKN}             \\ \hline
         1K  & 90.82$\pm$2.31 & 91.15$\pm$2.37 & 91.49$\pm$1.66             & 91.84$\pm$1.44           & \textbf{92.02$\pm$1.54}   \\
         2K  & 93.04$\pm$1.35 & 94.15$\pm$1.44 & 94.11$\pm$1.08             & 94.32$\pm$1.67           & \textbf{95.41$\pm$1.37}   \\
         5K  & 96.64$\pm$1.65 & 96.13$\pm$1.68 & 96.83$\pm$1.33             & \underline{98.45$\pm$1.58}           & \textbf{98.47$\pm$1.73}   \\
         10K & 98.79$\pm$1.14 & 94.81$\pm$1.07 & 98.80$\pm$0.86             & \underline{99.02$\pm$0.78}           & \textbf{99.03$\pm$0.74}   \\
         20K & 99.03$\pm$0.61 & 97.39$\pm$0.72 & 98.97$\pm$0.49             & 99.19$\pm$0.68           & \textbf{99.26$\pm$0.51}   \\
         40K & 99.25$\pm$0.53 & 98.21$\pm$0.49 & 99.10$\pm$0.53             & 99.27$\pm$0.61           & \textbf{99.32$\pm$0.27}   \\
         60K & 99.30$\pm$0.41 & 98.45$\pm$0.51 & 99.34$\pm$0.37             & 99.39$\pm$0.24 & \textbf{99.45$\pm$0.18}   \\
        \bottomrule
    \end{tabular}
    \caption{
        \normalsize  Classification accuracy (\%) for compared methods on the MNIST dataset without data augmentation. Here, we bold the optimal results and underline the results which show no significant difference with the optimal one.}
    \label{tab.mnnist}
\end{table*} 

\section{Experiments}
In this section, compared with related algorithms, we study the experimental performance of \texttt{CSKN} on several benchmark datasets to demonstrate the effects of factors: 
1) non-stationary spectral kernel, 2) deep architecture, 3) convolutional filters, 4) kernel learning via backpropagation.
We first run algorithms on five structural datasets with a small size. 
Then, for a medium-size dataset MNIST, we conduct experiments on different data partitions with varying sizes.

\subsection{Experimental Setup}
We use a three-layer network with $2000 \times 2000 \times 2000$ width for deep architectures to achieve favorable approximation for Monte Carlo sampling.
All algorithms are initialized according to \eqref{def.initial-spectral-density}, where the spectral density for $l$-th layer $s_l(\oo, \oo')$ is fixed on the critical line between ordered and chaotic phases according to mean field theory \cite{poole2016exponential,schoenholz2017deep}.
Specifically, convolutional networks use Delta-orthogonal initialization \cite{xiao2018dynamical}.
Using $5$-folds cross-validation, we select regularization parameters $\lambda_1, \lambda_2 \in \{10^{-10}, 10^{-9}, \cdots, 10^{-1}\}$.
We implement all algorithms using Pytorch \cite{paszke2019pytorch} and exert Adam as optimizer \cite{kingma2014adam} with the $32$ samples in a batch.
All experiments are repeated 10 times to obtain stable results, meanwhile those multiple test errors provide the statistical significance of the difference between compared methods and the optimal one. 
We make use of $2D$ convolutional filters on all datasets where the convolutional filters are $2 \times 2$ for the first layer while $3 \times 3$ for higher layers.

To confirm the effectiveness of factors used in our algorithm, we compare \texttt{CSKN} with several relevant algorithms:
1) \textbf{CNN}: Vanilla convolutional network only consists of convolutional layers (ReLU as activation) but without pooling operators and skip connections \cite{xiao2018dynamical}.\\
2) \textbf{CRFF}: Stacked random Fourier features \cite{zhang2017stacked} with convolutional filters, corresponding to stationary spectral kernels.\\
3) \textbf{DSKN}: Deep spectral kernel network without convolutional filters \cite{xue2019deep}. \\
4) \textbf{CDSK}: A variant of \texttt{CSKN} where hyparameters are just assigned and backpropagation is not used.

\subsection{Experiments on Small Image datasets}
We first run experiments on several small size image datasets where the structural information is more likely to be captured by convolution operators. 
These images datasets are collected in LIBSVM Data \cite{chang2011libsvm}.
We use the primal partition of training and testing data. 

We report the results in Table \ref{tab.small-dataset} where the results indicate: 
1) the proposed \texttt{CSKN} achieves optimal accuracies on all datasets, validating the effectiveness of our learning framework.
2) The results of CDSK are slightly worse than \texttt{CSKN} due to the lack of updates on parameters.
3) Compared with \texttt{CSKN}, DSKN shows poor performance because DSKN is a fully connected network without convolutional filters.
4) CRFF provides the worst results that coincide with the generalization analysis where stationary spectral kernel leads to inferior generalization error bounds.

\subsection{Handwriting recognition on MNIST}
Here, we conduct experiments on the MNIST dataset \cite{lecun1998gradient} which consists of 60,000 training images and 10,000 testings of handwritten digits. 
We randomly select a part of instances from training data to evaluate the performance on different partitions.

Test accuracies are reported in Table \ref{tab.mnnist}. 
The results illustrate: 
1) \texttt{CSKN} outperforms compared methods on all data size. 
2) Non-stationary kernels always provide better results than the stationary kernels approach (CRFF).
3) With appropriate initialization, even without backpropagation, convolutional deep spectral kernel (CDSK) can still achieve similar performance as \texttt{CSKN}.
4) Kernel-based networks work better than CNN on a small number of training samples.

\section{Conclusion and Discussion}
In this paper, we first integrate the non-stationary spectral kernel with deep convolutional neural network architecture, using Monte Carlo approximation for each layer.
The proposed algorithm is a trainable network where it optimizes the spectral density and the estimator together via backpropagation.
Then, based on Rademacher complexity, we extend the generalization analysis of kernel methods to the proposed network.
From the perspective of generalization, we prove non-stationary spectral kernel characterizes better generalization ability and deeper architectures lead to sharper error bounds with suitable initialization.
Generalization analysis interprets the superiority of deep architectures and can be applied to general DNN to improve their interpretability.
Intuitively, the generating feature mappings enjoy the following benefits: 1) \textit{input-dependent} (non-stationary spectral kernels), 2) \textit{output-dependent} (backpropagation towards the objective), 3) \textit{hierarchical represented} (deep architecture), 4) \textit{local correlated} (convolutional operators).

However, there are still a few tackle problems to be settled. For convolutional networks, current theoretical work focus on group invariance \cite{mallat2012group}, stability \cite{bietti2019group} and approximation ability \cite{zhou2020universality}. 
However, these theories can not explain why convolutional architectures work better than fully connected networks.
In future work, we will try to explain the generalization ability of convolutional networks using downsampling \cite{zhou2020theory} and locality.
Besides, generalization analysis indicates that the initialization variance $\sigma_l$ should be increased for the growth of depth, while $\sigma_l$ decreases as $l$ increasing in current mean field theory work \cite{schoenholz2017deep,xiao2018dynamical}.
It's worthy to explore the tradeoffs between generalization and optimization in terms of random initialization.
Our work also can be incorporated with \textit{Neural Tangent Kernel} (NTK) \cite{jacot2018neural} to capture the dynamics of signals and conduct a simpler kernel.

\section{Proof}
\begin{proof}[proof of Lemma \ref{lem.excess_risk_bounds}]
    Based on the $L$-Lipschitz condition, we combine Lemma A.5 of \cite{bartlett2005local} with the contraction lemma (Lemma 5 of \cite{cortes2016structured}).
    Then, with a probability at least $1-\delta$, there holds 
    \begin{align}
        \label{inequation.grc-bound-empirical-grc}
        \mathcal{E}(\widehat{f}_n) - \mathcal{E}(f^*) \leq 4 \sqrt{2} L \widehat{\mathcal{R}}(H_\kappa) + \mathcal{O}\Big(\sqrt{\frac{\log 1/\delta}{n}}\Big).
    \end{align}
    We estimate empirical Rademacher complexity via
    \begin{equation}
        \label{equation.grc-bound-grc-estimate}
        \begin{aligned}
            \widehat{\mathcal{R}}(H_\kappa) & = \frac{1}{n} ~ \mathbb{E}_{\boldsymbol \xi} \left[\sup_{f \in H_\kappa} \sum_{i=1}^n \sum_{k=1}^K \xi_{ik} [f(\xx_i)]_k  \right]             \\
                                               & = \frac{1}{n} ~ \mathbb{E}_{\boldsymbol \xi} \left[\sup_{f \in H_\kappa} \langle \WW, \boldsymbol{\Phi}_{\boldsymbol \xi} \rangle \right],
        \end{aligned}
    \end{equation}
    where $\WW, \boldsymbol{\Phi}_{\boldsymbol \xi} \in \mathcal{H} \times \mathbb{R}^{K}$ and $\langle \WW, \boldsymbol{\Phi}_{\boldsymbol \xi} \rangle = \text{Tr}(\WW^T\boldsymbol{\Phi}_{\boldsymbol \xi})$ and the matrix $\boldsymbol{\Phi}_{\boldsymbol \xi}$ is defined as follows:
    \begin{align*}
        \boldsymbol{\Phi}_{\boldsymbol \xi} := \left[\sum_{i = 1}^n \xi_{i1}\phi(\xx_i),
            \sum_{i = 1}^n \xi_{i2}\phi(\xx_i),
            \cdots, \sum_{i = 1}^n \xi_{iK}\phi(\xx_i)\right].
    \end{align*}
    Applying H\"older's inequality and $\|\WW\|_*$ bounded by a constant $B$ to (\ref{equation.grc-bound-grc-estimate}), there holds
    \begin{equation}
        \label{inequation.grc-bound-holder}
        \begin{aligned}
             & \widehat{\mathcal{R}}(H_\kappa)
            = ~   \frac{1}{n} ~ \mathbb{E}_{\boldsymbol \xi} \left[\sup_{f \in H_\kappa} \langle \WW, \boldsymbol{\Phi}_{\boldsymbol \xi} \rangle \right] \\
            &\leq ~ \frac{1}{n} ~ \mathbb{E}_{\boldsymbol \xi} \left[\sup_{f \in H_\kappa} \|\WW\|_* \|\boldsymbol{\Phi}_{\boldsymbol \xi}\|_F \right]        
             \leq \frac{B}{n} ~ \mathbb{E}_{\boldsymbol \xi} \left[\|\boldsymbol{\Phi}_{\boldsymbol \xi}\|_F \right] \\
            &\leq ~ \frac{B}{n} ~ \mathbb{E}_{\boldsymbol \xi} \left[\sqrt{\|\boldsymbol{\Phi}_{\boldsymbol \xi}\|_F^2} \right]
            \leq \frac{B}{n} ~ \sqrt{\mathbb{E}_{\boldsymbol \xi} ~ \|\boldsymbol{\Phi}_{\boldsymbol \xi}\|_F^2}.
        \end{aligned}
    \end{equation}
    Then, we bound $\mathbb{E}_{\boldsymbol \xi} ~ \|\boldsymbol{\Phi}_{\boldsymbol \xi}\|_F^2$ as follows
    \begin{equation}
        \label{inequation.grc-bound-estimate-Phi}
        \begin{aligned}
             & \mathbb{E}_{\boldsymbol \xi} ~ \|\boldsymbol{\Phi}_{\boldsymbol \xi}\|_F^2
            \leq ~  \mathbb{E}_{\boldsymbol \xi} ~ \sum_{k=1}^K \Big\|\sum_{i=1}^n \xi_{ik}\phi(\xx_i)\Big\|_2^2                                          \\
             & \leq ~  \sum_{k=1}^K \mathbb{E}_{\boldsymbol \xi} ~ \Big\|\sum_{i=1}^n \xi_{ik}\phi(\xx_i)\Big\|_2^2                                       \\
             & \leq ~ \sum_{k=1}^K \mathbb{E}_{\boldsymbol \xi} ~ \sum_{i,k=1}^n \xi_{ik}\xi_{jk} \big[\langle \phi(\xx_i), \phi(\xx_j) \rangle\big] \\
             & = ~    K \sum_{i=1}^n \langle \phi(\xx_i), \phi(\xx_i) \rangle.
        \end{aligned}
    \end{equation}
    The last step is due to the symmetry of the kernel $\kappa(\xx, \xx') = \langle \phi(\xx_i), \phi(\xx_i) \rangle$ .
    We finally bound the empirical Rademacher complexity
    \begin{equation}
        \label{inequation.grc-bound-proof-rff}
        \begin{aligned}
            \widehat{\mathcal{R}}(H_\kappa)
            \leq ~ & \frac{B}{n} \sqrt{K \sum_{i=1}^n \langle \phi(\xx_i), \phi(\xx_i) \rangle}
        \end{aligned}
    \end{equation}
    where $B = \sup_{f \in H_\kappa} \|\WW\|_*$.
    Substituting the above inequation (\ref{inequation.grc-bound-proof-rff})  to (\ref{inequation.grc-bound-empirical-grc}), we complete the proof.
\end{proof}


\bibliography{all}
\bibliographystyle{icml2020}

\end{document}